\documentclass[10pt,twocolumn,letterpaper]{article}

\usepackage[pagenumbers]{cvpr} 
\usepackage{graphicx}
\usepackage{float}
\usepackage{amsmath}
\usepackage{mathrsfs}
\usepackage{soul}
\usepackage{amssymb}
\usepackage{bbold}
\usepackage{tabularray}
\usepackage{multirow}
\usepackage{lineno}

\usepackage{microtype}

\usepackage[utf8]{inputenc}
\DeclareUnicodeCharacter{02CB}{`}

\usepackage{cuted}
\usepackage{graphicx} 

\definecolor{cvprblue}{rgb}{0.21,0.49,0.74}
\definecolor{lightpink}{HTML}{000B58}
\definecolor{mytable}{HTML}{DFF2EB}
\definecolor{customblue}{HTML}{789DBC}
\definecolor{customgreen}{HTML}{88C273}
\usepackage[pagebackref,breaklinks,colorlinks,allcolors=cvprblue]{hyperref}

\def\paperID{3344} 
\def\confName{CVPR}
\def\confYear{2025}

\title{Towards Zero-Shot Anomaly Detection and Reasoning \\ with Multimodal Large Language Models}

\author{Jiacong Xu$^{1}$\thanks{Most of this work was done when J. Xu was an intern at HRI-USA.} \hspace{0.1cm} Shao-Yuan Lo$^{2}$ \hspace{0.1cm} Bardia Safaei$^{1}$ \hspace{0.1cm} Vishal M. Patel$^{1}$ \hspace{0.1cm} Isht Dwivedi$^{2}$ \\
$^{1}$Johns Hopkins University \hspace{0.1cm} $^{2}$Honda Research Institute USA \\
{\tt\small \{jxu155, bsafaei1, vpatel36\}@jhu.edu \hspace{0.1cm} \{shao-yuan\_lo, idwivedi\}@honda-ri.com}
}

\begin{document}
\maketitle
\begin{abstract}
Zero-Shot Anomaly Detection (ZSAD) is an emerging AD paradigm. Unlike the traditional unsupervised AD setting that requires a large number of normal samples to train a model, ZSAD is more practical for handling data-restricted real-world scenarios. Recently, Multimodal Large Language Models (MLLMs) have shown revolutionary reasoning capabilities in various vision tasks. However, the reasoning of image abnormalities remains underexplored due to the lack of corresponding datasets and benchmarks. To facilitate research in AD \& reasoning, we establish the first visual instruction tuning dataset, \textbf{Anomaly-Instruct-125k}, and the evaluation benchmark, \textbf{VisA-D\&R}. Through investigation with our benchmark, we reveal that current MLLMs like GPT-4o cannot accurately detect and describe fine-grained anomalous details in images. To address this, we propose Anomaly-OneVision (\textbf{Anomaly-OV}), the first specialist visual assistant for ZSAD and reasoning. Inspired by human behavior in visual inspection, Anomaly-OV leverages a Look-Twice Feature Matching (LTFM) mechanism to adaptively select and emphasize abnormal visual tokens. Extensive experiments demonstrate that Anomaly-OV achieves significant improvements over advanced generalist models in both detection and reasoning. Extensions to medical and 3D AD are provided for future study. The link to our project page: \url{https://xujiacong.github.io/Anomaly-OV/}

\end{abstract}
    
\section{Introduction}
\begin{figure}[t]
\centering
    \includegraphics[width=\linewidth]{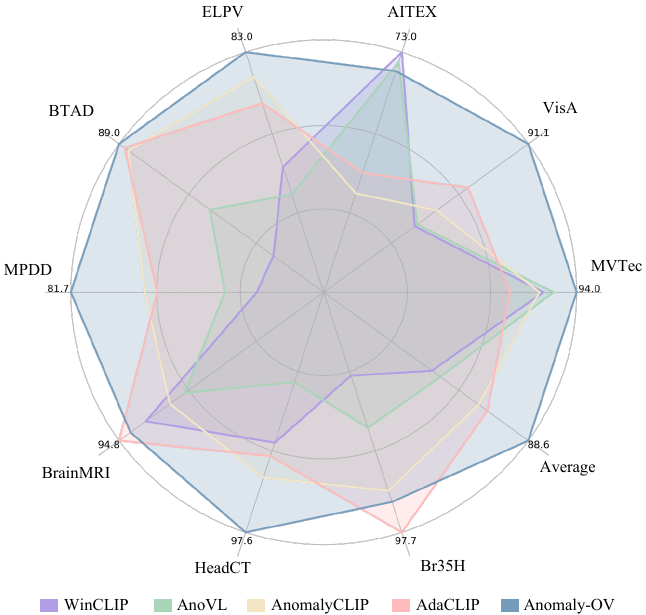}
\caption{Visualization of the image-level AUROC comparison between our Anomaly-OV and current state-of-the-art ZSAD methods (WinCLIP \cite{jeong2023winclip}, AnoVL \cite{deng2023anovl}, AnomalyCLIP \cite{zhou2024anomalyclip}, AdaCLIP \cite{cao2025adaclip}). Notably, our zero-shot performance on VisA even surpasses most recent advances in the few-shot setting \cite{li2024promptad, zhu2024toward, gu2024anomalygpt}.}
\label{fig:teaser}
 \vspace{-1mm}
\end{figure}

Visual Anomaly Detection (AD) is a well-established task in computer vision, extensively applied in scenarios such as industrial defect inspection \cite{mvtec, xie2023pushing, roth2022towards, huang2022registration, mou2023rgi, chen2022deep, bergmann2020uninformed, reiss2023mean, you2022a, cao2023anomaly} and medical image diagnosis \cite{wolleb2022diffusion, bmad, han2021madgan, huang2024adapting, zhang2024mediclip, wei2018anomaly, fernando2021deep, zhao2021anomaly}. In the traditional unsupervised AD (a.k.a. one-class AD) setting,  models learn the distribution of normal visual features from normal samples and are required to identify anomaly samples during inference. While recent advancements \cite{isaac2024towards, strater2024generalad, chen2024unified, tang2025incremental, he2024learning, yao2024glad, zhang2024realnet, lee2024text, yao2024hierarchical, ho2024long, fuvcka2025transfusion, hou2021divide} have significantly improved the detection performance, these approaches assume the availability of a substantial number of normal samples. However, this assumption becomes impractical in certain scenarios due to strict data privacy policies and the significant human effort required for data classification, sometimes involving experts or specialists. Therefore, Zero-Shot Anomaly Detection (ZSAD) is emerging as a popular research direction, leading to the development of many innovative methods \cite{jeong2023winclip, zhou2024anomalyclip, cao2025adaclip, li2024zero, li2024promptad2, gu2024filo, sato2023prompt, deng2024simclip, schwartz2024maeday, zhu2024llms}. 

\begin{figure}[t]
\centering
    \includegraphics[width=\linewidth]{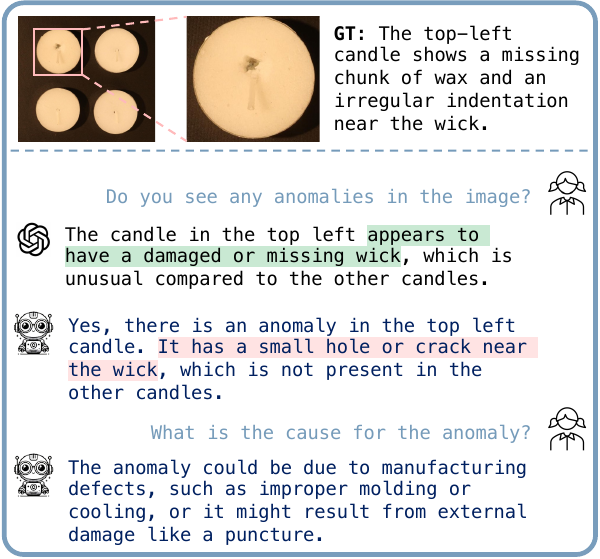}
\caption{Industrial image anomaly reasoning results from GPT-4o \cite{gpt-api-4o} and our Anomaly-OV. The responses for fine-grained anomaly reasoning are highlighted, with the ground truth given for reference.}
\label{fig:teaser2}
 \vspace{-2mm}
\end{figure}

Recent advances in Multimodal Large Language Models (MLLMs) \cite{blip, zhu2024minigpt, instructblip, chen2024spatialvlm, llava, llava15, llava_interleave, llavaonevision, llavamed} have shown revolutionary reasoning capabilities in various vision tasks \cite{lv2024video, yang2024follow, cheng2024emotion, guo2024stimuvar, zhou2024vicor, sermanet2024robovqa, xie2025funqa, zhou2024navgpt, nie2025reason2drive}. However, the reasoning of image abnormalities has not been explored due to the challenges of collecting large-scale datasets and establishing benchmarks. Existing methods simply predict the likelihood of an anomaly without providing rationales \cite{jeong2023winclip, zhou2024anomalyclip, cao2025adaclip, deng2023anovl, april-gan}. In contrast, for better interpretability, robustness, and trustworthiness, people would expect models to explain \textbf{why an image is considered anomalous} and provide visual evidence. Interestingly, we find that recent advanced MLLMs, such as GPT-4o \cite{gpt-api-4o}, fall short in AD \& reasoning. As shown in Figure \ref{fig:teaser2}, while the detection is correct, the explanation from GPT-4o lacks accuracy, indicating a gap in a comprehensive understanding of the anomaly.

To expedite research in AD \& reasoning, we establish the first visual instruction tuning dataset, Anomaly-Instruct-125k, and the evaluation benchmark, VisA-D\&R, through intensive human efforts. After evaluating current generalist MLLMs, we observe that these models fail to accurately detect and describe fine-grained anomalous details in images. To address this, we propose Anomaly-OneVision (Anomaly-OV), the first specialist visual assistant for ZSAD and reasoning. Unlike existing ZSAD methods \cite{jeong2023winclip, zhou2024anomalyclip, cao2025adaclip, deng2023anovl, april-gan}, Anomaly-OV directly learns object-awareness abnormality embeddings in feature space using only the visual encoder. Inspired by human behavior in visual inspection, Anomaly-OV employs a Look-Twice Feature Matching (LTFM) mechanism to assist its LLM in adaptively selecting and emphasizing the most suspicious abnormal visual tokens.

Extensive experiments demonstrate that Anomaly-OV achieves significant improvements over advanced generalist models in both detection and reasoning. Extended results of Anomaly-OV, from applications in industrial defect detection to 3D inspection and medical image diagnosis, are provided for future study. With precise descriptions and rationales of visual anomalies, our model can infer potential causes (see Figure \ref{fig:teaser2}), assess current impacts, and offer improvement suggestions, positioning itself as a reliable assistant for visual inspection. Our contributions are in two folds:
\begin{itemize}
	\item We establish the first visual instruction tuning dataset and benchmark for anomaly detection and reasoning.
	\item We propose the first specialist visual assistant with state-of-the-art performance for this new impactful domain.
\end{itemize}

\section{Related Work}

\noindent \textbf{Multimodal Large Language Models.}
Vision-Language Models (VLMs), such as CLIP \cite{clip}, exhibit robust zero-shot classification capabilities and have been applied to a range of downstream vision tasks \cite{lin2023clip, liang2023open, chen2023clip2scene, liu2023clip, wang2023clipn}. Combining a VLM's vision encoder and an LLM \cite{bert, roberta, t5}, MLLMs \cite{blip, blip2, llava, zhu2024minigpt, instructblip} enable text-based interactions related to visual content. MLLMs have shown remarkable reasoning capability, particularly when incorporated with prompting strategies such as Chain-of-Thought~\cite{wei2022chain, lu2022learn, zhang2024multimodal}. Recent studies have harnessed MLLMs to provide reasoning for downstream tasks, e.g., video anomaly detection~\cite{lv2024video,yang2024follow}, affective computing~\cite{cheng2024emotion,guo2024stimuvar}, and visual commonsense reasoning~\cite{zhou2024vicor}, revealing more interpretability.

\medskip
\noindent \textbf{Unsupervised Anomaly Detection.}
Due to the scarcity and difficulty of collecting anomalous data, researchers often focus on the unsupervised AD setting, which exclusively uses normal data to train an AD model. Earlier studies, such as reconstruction-based~\cite{lo2022adversarially,mou2023rgi,zavrtanik2021draem}, student-teacher~\cite{deng2022anomaly,tien2023revisiting,zhang2023destseg}, and augmentation-based~\cite{li2021cutpaste} approaches, assume a large amount of normal data is available. These traditional approaches are less practical when data are restricted or expensive, such as in the medical domain.

\medskip
\noindent \textbf{Zero-Shot Anomaly Detection.} Unlike unsupervised AD \cite{roth2022towards, huang2022registration} and few-shot AD \cite{zhu2024toward, li2024promptad, gu2024anomalygpt, fang2023fastrecon, huang2024adapting}, ZSAD models directly access the likelihood of abnormality for a given image without requiring data specific to the target object. Existing works \cite{jeong2023winclip, zhou2024anomalyclip, cao2025adaclip, deng2023anovl} accomplish ZSAD by comparing visual and textual features encoded by visual and text encoders of CLIP and constructing their positive (anomaly) and negative (normal) prompts in the format of:
\begin{align*}
&\mathcal{P}^{+}=[V_{1}][V_{2}]...[V_{n}][object] \\
&\mathcal{P}^{-}=[W_{1}][W_{2}]...[W_{n}][object]
\end{align*}
where $V_{i}$ and $W_{i}$ are handcrafted or learnable tokens, and $object$ refers to the word \texttt{object} or the class name of the object. However, simply utilizing \texttt{object} to represent all kinds of objects cannot capture the class-awareness abnormality types. Also, for an intelligent visual assistant, the images should be totally blind to the user (object-agnostic). 

\begin{figure*}[t]
\centering
    \includegraphics[width=\linewidth]{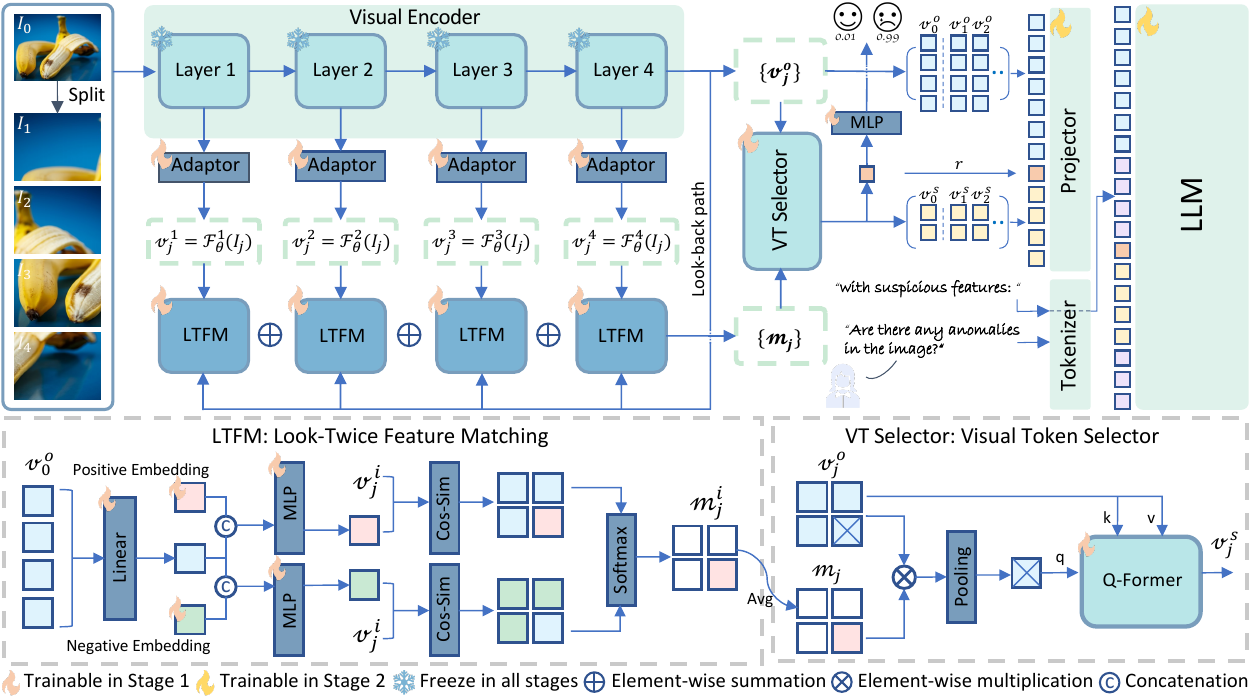}
\caption{Overview of the \textbf{Anomaly-OV} architecture. It consists of two training stages: (1) professional training for the anomaly expert, and (2) visual instruction tuning for anomaly detection and reasoning. Text and visual tokens are distinguished by different colors.}
\label{fig:main}
\end{figure*}

\section{Method}

\subsection{Preliminary}
Training an MLLM from scratch demands extensive data and computational resources to align the visual and textual embedding spaces and develop robust instruction-following capabilities. Recent studies \cite{xie2024emovit, face, pose} reveal that pre-trained MLLMs function as generalists, possessing a broad knowledge base but underperforming in specialized domains. Therefore, our goal is to introduce an auxiliary specialist or expert model designed to guide the generalist in selecting and utilizing critical visual tokens. This approach circumvents the need for large-scale pre-training while preserving the generalization capacity of the original model.

We choose \textit{LLaVA-OneVision} \cite{llavaonevision} as our base MLLM because it is open-sourced and performs similarly to other commercial models. \textit{LLaVA-OneVision} follows the model architectures for LLaVA family \cite{llava,llava15,llava_interleave,llavanext} and other generic MLLMs, which typically consist of three major components: Visual Encoder, Projector, and LLM. The visual encoder \cite{clip, siglip} extracts the visual information from the raw images, the projector aligns the spaces of visual features with the word embedding, and the LLM is responsible for textual instruction processing and complex reasoning. Since the image resolution for CLIP pre-training is fixed, \textit{LLaVA-OneVision} leverages AnyRes with pooling strategy to scale up the input raw image resolution. Specifically, the high-resolution images are divided into a prototyped number of crops, and the visual encoder independently processes the image crops before final spatial pooling.

\subsection{Architecture Overview}
With the same image-splitting strategy \textit{AnyRes} as \textit{LLaVA-OneVision}, the input high-resolution image is split into several crops, and the new image set can be written as:
\begin{equation}
    \mathcal{I}=\{I_{0}, I_{1}, I_{2}, ..., I_{n-1}\}
\end{equation}
where $I_{0}$ is the resized original image and $I_{j\neq 0}$ refers to the image crops. As shown in Figure \ref{fig:main}, the image set $\mathcal{I}$ will be processed by the visual encoder $\mathcal{F}_{\theta}$ to generate the final visual features $\{\mathbf{v}^{o}_{j}\}$. Similar to AnomalyCLIP \cite{zhou2024anomalyclip}, we store the outputs for four selected layers in the ViT \cite{vit} to capture the image representations from different levels and apply four adapters to compress the feature dimension. Then, the extracted visual features can be written as:
\begin{equation}
    \mathbf{v}^{i}_{j} = \mathcal{F}^{i}_{\theta}(I_{j})
\end{equation}
where $i$ denotes the $i$-th level and $j$ refers to the index of corresponding image in $\mathcal{I}$. These multi-level features have been demonstrated to be effective in capturing fine-grained local semantics by recent works \cite{gu2024anomalygpt, cao2025adaclip, zhou2024anomalyclip}.

The large-scale pre-trained CLIP models align the projection spaces of the textual and visual encoder. Therefore, the encoded image features already contain the class information required by ZSAD. To avoid human involvement in object classification and reduce the model complexity, we remove the heavy text encoder commonly utilized in existing works and let the visual model itself parse the information for suspicious classes or objects. Specifically, the output visual features for the original image $\mathbf{v}^{o}_{0}$ are leveraged to provide the global description of the target object or regions in the look-back path. With the multi-level features and the global embeddings, the LTFM module is responsible for the recognition and localization of suspicious tokens.

\begin{figure}[t]
\centering
    \includegraphics[width=\linewidth]{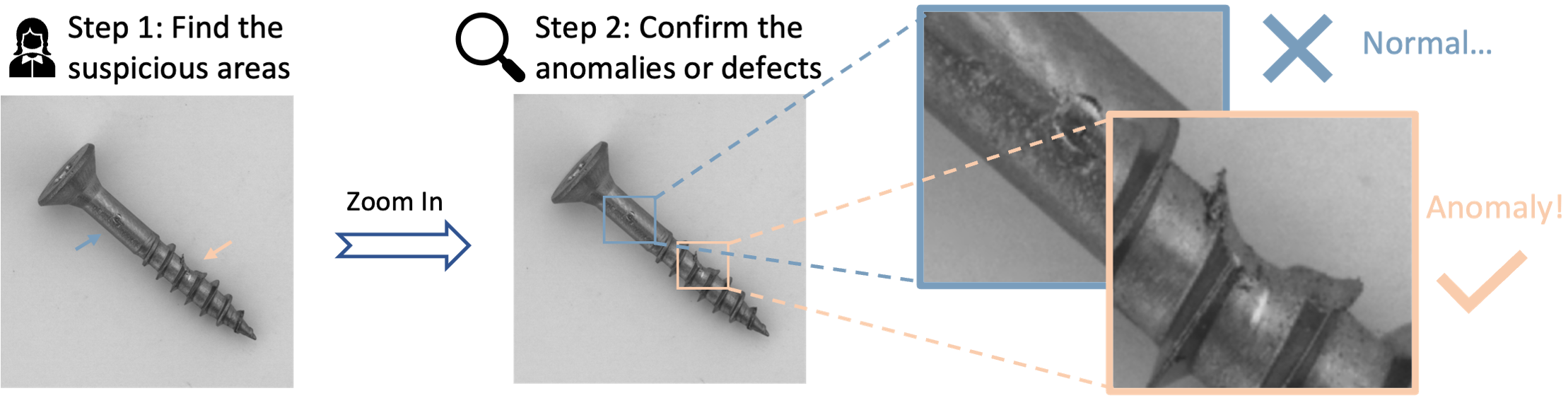}
\caption{Simulation of visual anomaly inspection by humans.}
\label{fig:human}
\vspace{-4mm}
\end{figure}
Drawing inspiration from human visual inspection, where suspicious objects or regions are identified and then inspected closely (see Figure \ref{fig:human}), we design the VT selector module for aggregating (zooming in) the crucial visual tokens and explicitly assisting the LLM in distinguishing these tokens from many irrelevant ones when dealing with instructions regarding anomaly detection and reasoning. Additionally, the original visual features are preserved to maintain the generalization capability of the base model on regular instructions, such as \texttt{Can you describe the content of the image?}

\subsection{Look-Twice Feature Matching}
\label{ltfm}
Given the global object information $\mathbf{v}^{o}_{0}$ provided by the look-back path, we generate the class-awareness abnormality description by merging $\mathbf{v}^{o}_{0}$ with two learnable embeddings: $\mathbf{e}^{+}\in \mathbb{R}^{D}$ and $\mathbf{e}^{-}\in \mathbb{R}^{D}$, where $+$ and $-$ indicate positive (anomalous) and negative (normal) patterns and $D$ is the embedding dimension. Specifically, a linear layer $\mathcal{T}^{o}_{i}$ is applied along the token dimension to select and fuse useful tokens from $\mathbf{v}^{o}_{0}$, and then the fused vector will be concatenated with $\mathbf{e}^{+}$ and $\mathbf{e}^{-}$ independently and pass through two MLPs $\{\mathcal{G}^{+}_{i}, \mathcal{G}^{-}_{i}\}$ to generate the abnormality and normality descriptions $\{\mathbf{d}^{+}_{i}, \mathbf{d}^{-}_{i}\}$, which can be represented by:
\begin{align}
\{\mathbf{d}^{+}_{i}, \mathbf{d}^{-}_{i}\} = 
\begin{cases}
\mathcal{G}^{+}_{i}(\mathbf{e}^{+} \circ \mathcal{T}^{o}_{i}(\mathbf{v}^{o}_{0})) \\
\mathcal{G}^{-}_{i}(\mathbf{e}^{-} \circ \mathcal{T}^{o}_{i}(\mathbf{v}^{o}_{0}))
\end{cases}
\end{align}
The visual features extracted from different levels of the ViT focus on different scales of semantics. Thus, the parameters of $\mathcal{T}^{o}_{i}$ and $\{\mathcal{G}^{+}_{i}, \mathcal{G}^{-}_{i}\}$ should be independent for different levels, where $i$ indicate the level number. 

Similar to the zero-shot classification mechanism of CLIP models, we calculate the possibilities of each patch token in $\mathbf{v}^{i}_{j}$ belonging to the anomalous patterns by combining cosine similarity and softmax operations:
\begin{equation}
    \textbf{m}^{i}_{j}=\frac{exp(<\mathbf{d}^{+}_{i}, \mathbf{v}^{i}_{j}>/\tau)}{exp(<\mathbf{d}^{+}_{i}, \mathbf{v}^{i}_{j}>/\tau)+exp(<\mathbf{d}^{-}_{i}, \mathbf{v}^{i}_{j}>/\tau)}
\end{equation}
where $\textbf{m}^{i}_{j}$ represents the significance map for visual tokens, $\tau$ is the temperature hyperparameter, and $<,>$ refers to the cosine similarity operator. The patch weight in $\textbf{m}^{i}_{j}$ indicates the closeness of the corresponding visual token to the anomalous pattern. Then, all the maps are averaged to capture the token significances from low to high levels:
\begin{equation}
    \textbf{m}_{j}=\sum^{4}_{i=1}\textbf{m}^{i}_{j}/4
\end{equation}
The visual features are leveraged twice in the forward and look-back paths, so this module is named by \textit{Look-Twice Feature Matching} (LTFM), following the nature of two-step human visual inspection shown in Figure \ref{fig:human}.
\begin{figure*}[t]
\centering
    \includegraphics[width=\linewidth]{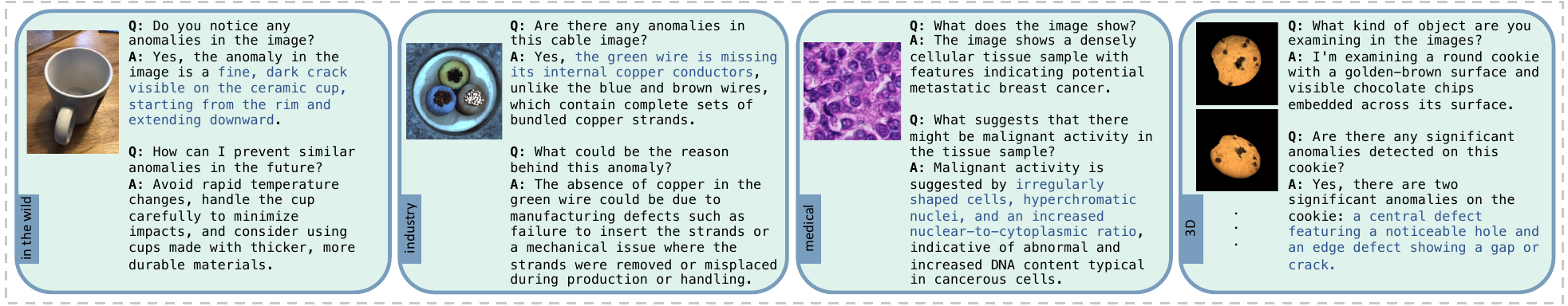}
\caption{Composition of the instruction data in \textbf{Anomaly-Instruct-125k}. There are four main types of image samples: \textit{in-the-wild}, \textit{industrial}, \textit{medical}, and \textit{3D} (in the format of multi-view images), covering most image anomaly detection tasks and enabling the possibility of a unified assistant for visual inspection. The reasoning words are highlighted in blue. For more information about dataset establishment, statistics, and the data collection pipeline, please refer to Section \ref{sup_dataset} in the supplementary.} 
\label{fig:daatset}
\vspace{-2mm}
\end{figure*}
\subsection{Visual Token Selector}
\label{vt_selector}
Under the image cropping strategy widely applied in recent MLLMs, there will be a large number of visual tokens for a high-resolution image, e.g., 7290 tokens for an image with 1152$\times$1152 resolution in \textit{LLaVA-OneVision}. While these tokens provide rich visual details, the LLM is required to pick the most useful information when adapting to a specific task. When the LLM lacks enough knowledge in this domain, the token-picking process will become complicated. Thus, our solution is to introduce a specialist or expert who knows which token is crucial or not and assist the LLM in selecting and emphasizing (zooming in) the crucial tokens.

Given the encoded visual tokens $\{\mathbf{v}^{o}_j\}$ for each image crop in $\mathcal{I}$ and the corresponding significance map $\textbf{m}_{j}$, the suspicious tokens are emphasized by direct multiplication of the two tensors. Then, the normal tokens will be scaled to zero while the anomalous tokens will be maintained. After that, spatial average pooling $\mathcal{P}$ is applied to reduce the number of tokens. This process can be written as:
\begin{equation}
    \textbf{q}_{j}=\mathcal{P}(\mathbf{v}^{o}_{j}\odot \textbf{m}_{j})
\end{equation}
where $\textbf{q}_{j}\in \mathbb{R}^{h \times w \times D}$ refers to the pooled query tokens. Empirically, setting $h=w=2$ provides a better trade-off than other options. Then, a Q-Former $\mathcal{Q}$ \cite{blip2} is leveraged to aggregate the correlated tokens in the original output by forwarding $\textbf{q}_{j}$ as the query and $\mathbf{v}^{o}_{j}$ as the key and value:
\begin{equation}
    \mathbf{v}^{s}_{j}=\mathcal{Q}(\textbf{q}_{j}, \mathbf{v}^{o}_{j}, \mathbf{v}^{o}_{j})
\end{equation}
The Visual Token Selector (VT Selector) serves as a tool for the anomaly expert to hand-pick visual tokens that contain the most suspicious semantics for a given image.

\subsection{Inference and Loss}
\noindent\textbf{Anomaly Prediction.} In the traditional anomaly detection task, the model predicts the possibility of the image being abnormal. To achieve anomaly score prediction, we aggregate the anomaly information from all the image crops by an average operation weighted on the significance maps:
\begin{equation}
    \mathbf{r}(\mathcal{I}) = \frac{\sum_{j, k} \mathbf{v}^{s}_{j}[k]\cdot \mathcal{P}(\textbf{m}_{j})[k]}{\sum_{j, k}\mathcal{P}(\textbf{m}_{j})[k]}
\end{equation}
where $\mathcal{P}$ is the same spatial pooling in VT Selector and $\mathbf{r}(\mathcal{I})$ is a vector containing the global anomaly information for the entire image. Then, the anomaly expert can calculate the image-level abnormal possibility by parsing $\mathbf{r}(\mathcal{I})$:
\begin{equation}
    \mathbf{score}(\mathcal{I})=Sigmoid(\mathcal{G}^{o}(\mathbf{r}(\mathcal{I})))
\end{equation}
where $\mathcal{G}^{o}$ is an MLP for distinguishing normal and abnormal semantics. To handle the unbalanced sample distribution, we employ the balanced BCE loss as the professional training objective for the anomaly expert components.

\medskip
\noindent\textbf{Text Generation.} Instead of directly forwarding the concatenation of the original $\{\mathbf{v}^{o}_{j}\}$ and the selected $\{\mathbf{r}(\mathcal{I}), \mathbf{v}^{s}_{j}\}$ visual tokens into the LLM, we apply an indication prompt \texttt{with <adv> suspicious feature:} in the middle of the two series of tokens, which will highlight the selected tokens for LLM when handling anomaly-related instructions. This approach can be considered a form of prompt engineering in MLLMs. Besides, the \texttt{<adv>} is chosen from \{\texttt{highly}, \texttt{moderately}, \texttt{slightly}\} and is determined by $\mathbf{score}(\mathcal{I})$ and predefined thresholds $\{\textbf{s}_{low}, \textbf{s}_{high}\}$. When the input image $\mathcal{I}$ has a high likelihood of anomaly, the LLM will place greater emphasis on the selected tokens; otherwise, these tokens will have less significance. The text generation is implemented by the original auto-regressive token prediction mechanism of LLM:
\begin{equation}
    p(X_{a}|\mathcal{I}, X_{q})=\prod^{L}_{t=1}p_{\theta}(x_{t}|\mathcal{I}, X_{q, <t}, X_{a, <t})
\end{equation}
where $X_{a, <t}$ and $X_{q, <t}$ are the answer and instruction tokens from all prior turns before the current prediction token $x_{t}$ for a sequence of length $L$. The entire model is parameterized by $\theta$ and trained by the original language model cross-entropy loss for each predicted answer token $x_{t}$.

\section{Dataset and Benchmark}
The lack of multimodal instruction-following data for image anomaly detection and reasoning hinders the development of special intelligent assistants in this domain. Even though AnomalyGPT \cite{gu2024anomalygpt} introduces a prompt tuning dataset by simulating the anomalies, the scale of their dataset and the diversity of their instructions and answers are limited, only focusing on anomaly localization. To resolve the data scarcity issue, we establish the first large-scale instruction tuning dataset: \textbf{Anomaly-Instruct-125k} and the corresponding anomaly detection and reasoning benchmark: \textbf{VisA-D\&R}.

\subsection{Anomaly-Instruct-125k}
LLaVA \cite{llava} builds its instruction tuning dataset by leveraging the image caption and bounding boxes available in the COCO dataset\cite{coco} to prompt the text-only GPT-4. ShareGPT4V \cite{sharegpt4v} provides a higher-quality dataset by directly prompting GPT-4V \cite{gpt-api-4vision}. However, there is no image caption provided in existing anomaly detection datasets \cite{mvtec, bmad}, and no matter GPT-4V \cite{gpt-api-4vision} or most recent GPT-4o  \cite{gpt-api-4o} cannot accurately locate and describe the anomalies in the image without explicit human involvement. 

To resolve these issues, we design a new prompt pipeline for accurate anomaly description generation. Since most of the datasets contain annotations for anomaly types, we manually combine the class name and anomaly type, such as \texttt{a [capsule] with [poke] on surface}. If the anomaly masks are provided, we draw bounding boxes on the images to highlight the anomalous area. The short description and the image with (or w/o) bounding boxes are used to prompt GPT-4o to generate the detailed image and anomaly descriptions. Then, we employ an in-context learning strategy similar to LLaVA to create the instructions.

For a unified visual inspection dataset, precise instruction data is collected from MVTec AD \cite{mvtec}, the training set of BMAD \cite{bmad}, Anomaly-ShapeNet \cite{anomaly_shapenet}, Real3D-AD \cite{real3d}, and MVTec-3D AD \cite{mvtec3d}, covering both 2D to 3D data across industry to medical domains. The 3D point cloud data are converted into 9 multi-view images, and the corresponding masks are rendered using predefined camera positions. However, the diversities and scales of these datasets are relatively limited, probably due to the difficulty of collecting anomaly images. To scale up the instruction data, we introduce an automatic anomaly data collection pipeline combining GPT-4o \cite{gpt-api-4o} and Google Image Search \cite{google-image-search} for image collection, data cleaning, and instruction generation. Finally, 72k in-the-wild images (named as WebAD) targeting anomaly detection are collected, significantly enriching our instruction dataset. Several samples from Anomaly-Instruct-125k are shown in Figure \ref{fig:daatset}. The instructions are mainly in the format of multi-round conversations, covering anomaly detection and description in low-level reasoning and potential cause and future suggestions for complex understanding. 

\subsection{VisA-D\&R}
VisA \cite{visa} is a classic but challenging industrial anomaly detection dataset, providing fine-grained anomaly type and segmentation for each image. For evaluation of the anomaly detection and reasoning performance on existing and future methods, we select 10 classes from VisA and follow a similar data generation pipeline of Anomaly-Instruct-125k to create the benchmark. Differently, significant human effort has been invested in meticulously reviewing all generated images and anomaly descriptions. Wrong descriptions are picked out and re-annotated by humans before utilizing them for Q\&A generation. Totally, the benchmark consists of 761 normal samples and 1000 anomalous ones. 

\begin{figure}[t]
\centering
    \includegraphics[width=\linewidth]{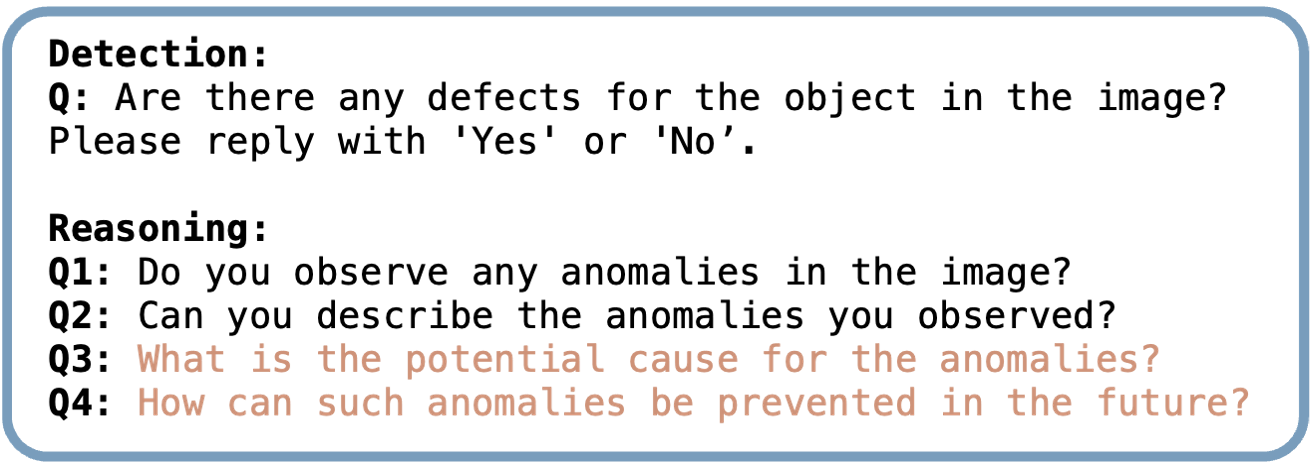}
\caption{Prompt examples in VisA-D\&R for detection and reasoning. The complex reasoning instructions are highlighted.} 
\label{fig:bench}
\vspace{-4mm}
\end{figure}

For evaluating detection performance, questions designed to elicit a one-word answer are used to prompt the MLLMs (Figure \ref{fig:bench}), with results quantified using Accuracy, Precision, Recall, and F1-score. We divide the reasoning performance into two parts: low-level reasoning that focuses on the description of visual defects or anomalies and complex reasoning requiring the MLLMs to provide the potential cause and future improvement strategies for the detected anomalies, where ROUGE-L \cite{rouge}, Sentence-BERT (SBERT) \cite{sbert}, and GPT-Score (GPT-4 as the judge \cite{llava}) are utilized to quantify the similarity between generated text and ground truth. Note that low-level reasoning is highly correlated to detection performance, while anomaly-type descriptions of low-level reasoning determine the output of complex reasoning.

\section{Experiments}
\subsection{Training \& Evaluation}
There are two independent training stages for Anomaly-OV. In Stage 1, the components of the anomaly expert are trained to obtain the token selection capability, targeting traditional ZSAD. This stage utilizes all of the data with anomaly labels in Anomaly-Instruct-125k. Similar to previous works \cite{zhou2024anomalyclip, cao2025adaclip}, when evaluating the model on the datasets contained in the training set, the corresponding datasets are replaced by VisA \cite{visa}. In Stage 2, the anomaly expert and visual encoder are frozen, while the projector and LLM are trainable. In addition to our instruction dataset, we sample around 350k data from the original training recipe of \textit{LLaVA-OneVision} to maintain the generalization ability. For more details on training, please refer to the supplementary.

The ZSAD performance for the anomaly expert is evaluated on nine benchmarks, including MVTec AD \cite{mvtec}, VisA \cite{visa}, AITEX \cite{aitex}, ELPV \cite{elpv}, BTAD \cite{btad}, and MPDD \cite{mpdd} for industrial inspection, and BrainMRI \cite{brainmri}, HeadCT \cite{headct}, and Br35H \cite{Br35h} for medical diagnosis. AUROC (Area Under the Receiver Operating Characteristic) is leveraged to quantify the image-level AD performance. For text-based anomaly detection, both normal and anomaly data are employed to assess the accuracy by examining if the generated text contains the word \texttt{Yes}. Differently, only anomaly data are utilized to prompt the MLLMs to determine their anomaly reasoning capabilities since the justifications of normality are similar for different models.

\begin{table}[ht]
\small
\centering
\begin{tabular}{lcccc}
\toprule
Method             & MVTec & VisA & HeadCT & BrainMRI \\
\hline
Full Model         &  \textbf{94.0}     &  \textbf{91.1}    &   \textbf{97.6}     &   93.9       \\
\hline
w/o. Look-back      & 92.8      & 90.5     &  96.6      &   93.5      \\
w/o. $\mathbf{e}^{+}$ \&  $\mathbf{e}^{-}$    &  92.1     & 90.1     & 94.7       &  92.9       \\
w/o. Q-Former       & 91.7      & 89.9     & 92.8       &  \textbf{95.1}       \\
w/o. WebAD          & 88.5      & 88.9     & 91.2       &  93.4       \\
\bottomrule
\end{tabular}
\caption{Ablation study for the anomaly expert of Anomaly-OV. w/o. Look-back refers to the removal of $\mathbf{v}^{o}_{0}$ in LTFM.}
\label{tab:2}
\vspace{-4mm}
\end{table}
\subsection{Zero-Shot Anomaly Detection}
\begin{table*}
\small
\centering
\begin{tabular}{lcccccccccc}
\toprule
\multicolumn{1}{l}{\multirow{2}{*}{Model}} & \multicolumn{6}{c}{Industrial Defects} & \multicolumn{3}{c}{Medical Anomalies} & \multicolumn{1}{c}{\multirow{2}{*}{Average}}  \\ 
\cmidrule(lr){2-7}\cmidrule(lr){8-10}
\multicolumn{1}{c}{}             & MVTec AD           & VisA & AITEX & ELPV & BTAD & MPDD & BrainMRI          & HeadCT & Br35H &  \multicolumn{1}{c}{}       \\
\hline
CLIP \cite{clip}        & 74.1               & 66.4 & 71.0  & 59.2 & 34.5 & 54.3 & 73.9              & 56.5   & 78.4  & 63.1    \\
CoOp \cite{coop}       & 88.8               & 62.8 & 66.2  & 73.0 & 66.8 & 55.1 & 61.3              & 78.4   & 86.0  & 70.9    \\
WinCLIP \cite{jeong2023winclip}     & 91.8               & 78.8 & \textbf{73.0}  & 74.0 & 68.2 & 63.6 & 92.6              & 90.0   & 80.5  & 79.2    \\
APRIL-GAN \cite{april-gan}  & 86.2               & 78.0 & 57.6  & 65.5 & 73.6 & 73.0 & 89.3              & 89.1   & 93.1  & 78.4    \\
AnoVL \cite{deng2023anovl}       & \underline{92.5}               & 79.2 & \underline{72.5}  & 70.6 & 80.3 & 68.9 & 88.7              & 81.6   & 88.4  & 80.3    \\
AnomalyCLIP \cite{zhou2024anomalyclip} & 91.5               & 82.1 & 62.2  & \underline{81.5} & 88.3 & \underline{77.0} & 90.3              & \underline{93.4}   & 94.6  & 84.5    \\
AdaCLIP \cite{cao2025adaclip}    & 89.2               & \underline{85.8} & 64.5  & 79.7 & \underline{88.6} & 76.0 & \textbf{94.8}              & 91.4   & \textbf{97.7}  & \underline{85.3}    \\
Ours  & \textbf{94.0}      & \textbf{91.1} & 72.0  & \textbf{83.0} & \textbf{89.0} & \textbf{81.7} & \underline{93.9}              & \textbf{97.6}   & \underline{95.5}  & \textbf{88.6}    \\
\bottomrule
\end{tabular}
\caption{Quantitative comparison of Image-level AUROC on different ZSAD methods (some of the results are borrowed from \cite{zhou2024anomalyclip, cao2025adaclip, zhu2024finegrainedabnormalitypromptlearning}). The best and the second-best results are bolded and underlined, respectively. Please refer to the supplementary for more detailed results.}
\label{Tab:1}
\vspace{-3mm}
\end{table*}
As shown in Table \ref{Tab:1}, compared with existing methods, the anomaly expert of Anomaly-OV achieves significant image-level AUROC improvements on most of the ZSAD benchmarks, which demonstrates that the text encoder widely applied in existing models is not necessary. The success of our model mainly originates from the extra data of WebAD (Table \ref{tab:2}), which enables the model to learn more generic semantics for normality and abnormality from the data distribution in the absence of the text encoder. This observation also reveals that large-scale in-the-wild online data can benefit zero-shot performance in anomaly detection. 

\begin{figure}[ht]
\centering
    \includegraphics[width=0.95\linewidth]{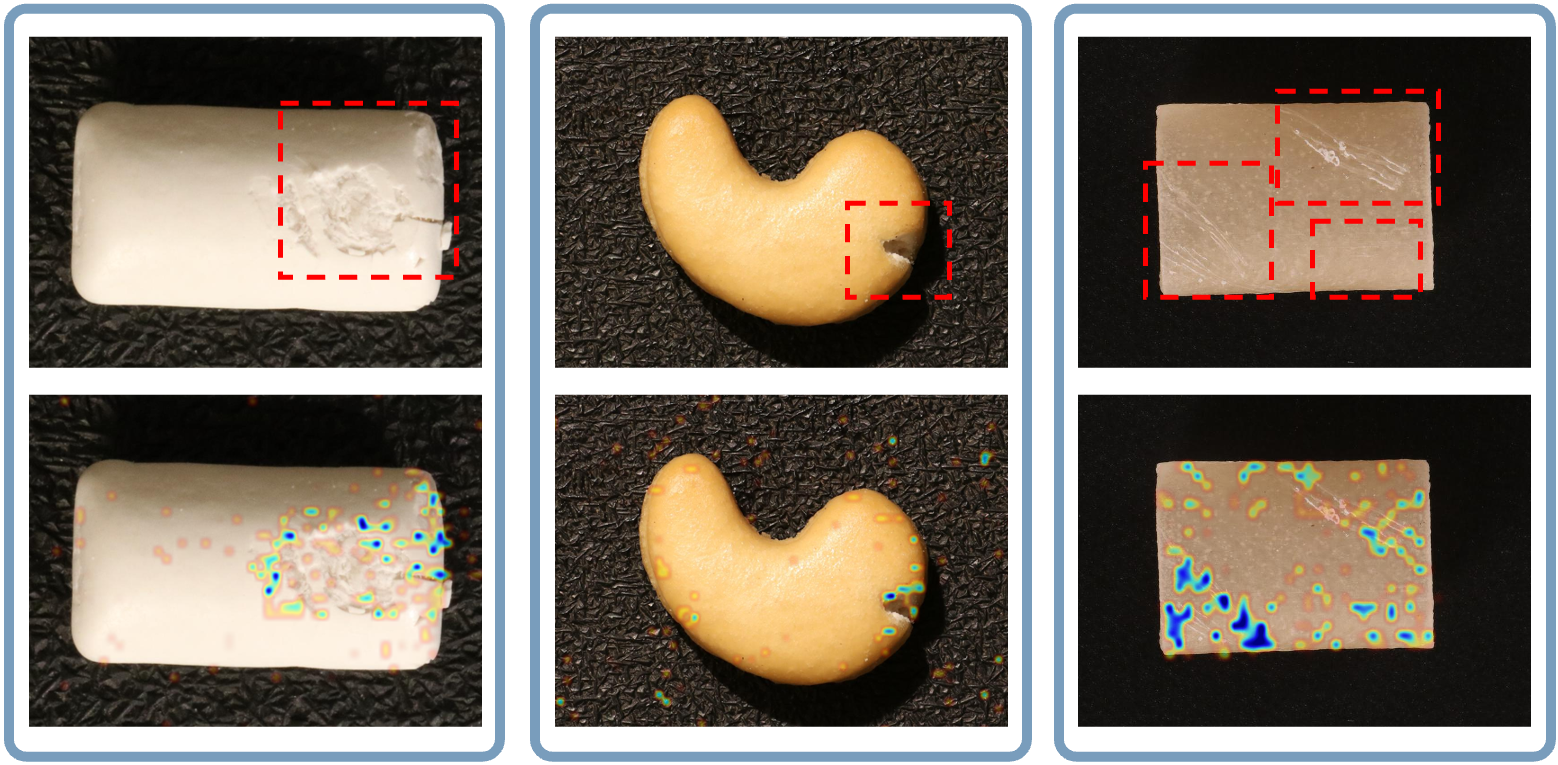}
\caption{Visualization of the significance map on VisA samples.} 
\label{fig:selection}
\vspace{-4mm}
\end{figure}

While the Q-Former reduces the model performance on BrainMRI, it shows effectiveness on most benchmarks, indicating the importance of token aggregation. Similarly, the look-back information and two learnable embeddings are required for describing class-awareness abnormality and distinguishing positive and negative features, respectively. As previously discussed, the anomaly expert is responsible for selecting suspicious visual tokens for the LLM, and the significance maps shown in Figure \ref{fig:selection} demonstrate the interpretable token selection mechanism. The high intensities are automatically distributed around the anomalous areas even without any supervision of the anomaly masks. 

\begin{table}
  \begin{minipage}{0.99\linewidth}
\centering
\scalebox{0.80}{
\begin{tabular}{l p{7.5cm} }
\toprule
 \multicolumn{2}{l}{\bf PCB Example: Excessive Solder}  \\
\midrule
&  \includegraphics[height=2.5cm]{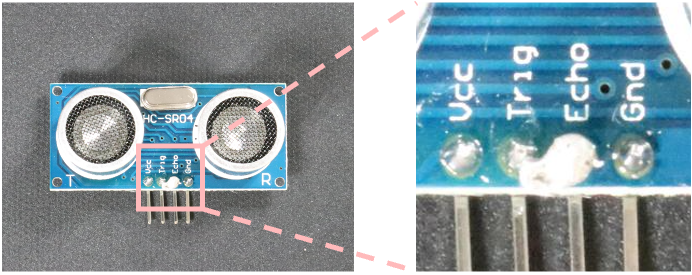} \\

User & Are there any anomalies visible in the image? \\
\midrule
GPT-4o & The image shows an HC-SR04 ultrasonic sensor. There don't appear to be any visible anomalies with the sensor itself. The components and pins seem intact and properly aligned. \\ \midrule
LLaVA-OV & The image shows an ultrasonic sensor module, specifically the HC-SR04 model. There are no visible anomalies in the image; it appears to be a standard representation of this type of sensor module. \\
\midrule
Ours & Yes, there is a noticeable anomaly in the image: \textcolor{customblue}{a short circuit on the PCB, indicated by a white, crusty residue near the 'Echo' pin.} \\
\bottomrule
\end{tabular}
}
\captionof{table}{Anomaly-OV presents more accurate anomaly detection.}
\label{tab:pcb2}  
  \end{minipage}
  \vspace{-4mm}
\end{table}

\subsection{Anomaly Detection \& Reasoning}
\begin{table*}
\small
\centering
\begin{tabular}{lccccccccc}
\toprule
\multicolumn{1}{l}{\multirow{2}{*}{Model}} & \multicolumn{4}{c}{Anomaly Detection} & \multicolumn{3}{c}{Low-level Reasoning} &  \multicolumn{2}{c}{Complex Reasoning}\\ 
\cmidrule(lr){2-5}\cmidrule(lr){6-8}\cmidrule(lr){9-10}
                 & Accuracy           & Precision & Recall & F1-score & ROUGE-L               & SBERT & GPT-Score & SBERT                & GPT-Score \\
\hline
GPT-4V \cite{gpt-api-4vision}          & 0.68              & 0.90      & 0.49   & 0.55     &        0.16             &   0.65    &      3.31     &  0.77                    &     5.64      \\
GPT-4o \cite{gpt-api-4o}          & 0.70              & 0.83      & 0.71   & 0.68     &           0.24          &    0.71   &    \textbf{4.84}       &  0.81                    &      \textbf{6.89}     \\
Qwen2-VL-2B \cite{qwen2vl}      & 0.65              & 0.87      & 0.55   & 0.59     & 0.22                    &   0.55    &  1.94         & 0.74                     &   4.26        \\
Qwen2-VL-7B \cite{qwen2vl}     & 0.76              & \underline{0.91}      & 0.69   & 0.75     & 0.25                    &  0.61     &  3.09         & 0.68                     &  4.62         \\
InternVL-2-8B \cite{internvl}   & 0.74              & 0.78      & 0.81   & 0.76     & 0.23                    & 0.73      &  3.69         & 0.80                     & 5.08          \\
InternVL-2-26B \cite{internvl}  & 0.73              & 0.86      & 0.66   & 0.68     & 0.21                    & \textbf{0.74}      &  4.13         & 0.80                     &   5.49        \\
IXC-2.5-7B \cite{ixc25}      & 0.72              & 0.88      & 0.63   & 0.67     & 0.21                    &  0.58     &  2.45         & 0.77                     &   5.14        \\
LLaVA-OV-0.5B \cite{llavaonevision}   & 0.54              & 0.70      & 0.19   & 0.28     & 0.20                    & 0.63      &  2.54         & 0.81                     &   4.34        \\
LLaVA-OV-7B \cite{llavaonevision}     & 0.71              & \textbf{0.95}      & 0.56   & 0.63     & 0.24                    & 0.66      &  3.57         &   0.79                   &  5.44         \\
\hline
LLaVA-OV-0.5B*   & 0.71              & 0.77      & \underline{0.84}   & 0.76     & 0.31                    &  0.70     & 3.69          & 0.82                     &  5.31         \\
Anomaly-OV-0.5B & \textbf{0.79}              & 0.86      & 0.83   & \underline{0.82}     & \underline{0.33}                    & 0.72      & 3.87          &  \underline{0.83}                    &  5.67         \\
Anomaly-OV-7B    & \textbf{0.79}                  & 0.83          & \textbf{0.86}      & \textbf{0.83}         &  \textbf{0.34}                   & \underline{0.73}      & \underline{4.26}          & \textbf{0.84}                     &  \underline{6.34}        \\
\bottomrule
\end{tabular}
\caption{Quantitative comparison of text-based anomaly detection and reasoning for MLLMs. Notably, the Accuracy and F1-score for the anomaly expert of Anomaly-OV can be calculated as $\{0.78, 0.77\}$ with threshold $0.5$. * indicates the model is fine-tuned on our dataset.}
\label{Tab:3}
 \vspace{-3mm}
\end{table*}
With the strong capabilities of the anomaly expert for zero-shot detection and suspicious token selection, Anomaly-OV accomplishes significant improvement in text-based anomaly detection and reasoning over other open-sourced generalist MLLMs, as shown in Table \ref{Tab:3}. Here are a few observations: \textit{i) While a larger language model cannot guarantee better detection performance, it always provides greater reasoning ability; ii) Most of the existing MLLMs present much lower recall than precision, indicating their insensitivity to visual anomalies; iii) GPT-4o shows stronger reasoning ability compared to other open-sourced models.} Table \ref{tab:pcb2} and Table \ref{tab:pasta} provide the qualitative comparison of our Anomaly-OV with its base model LLaVA-OV-7B \cite{llavaonevision} and GPT-4o \cite{gpt-api-4o}. Both GPT-4o and LLaVA-OV show insensitivity to anomalous features and cannot accurately detect the anomaly in the image. Sometimes, GPT-4o knows the image is anomalous but fails to describe the anomalies precisely.

We provide the fine-tuned version of the base model LLaVA-OV-0.5B on Anomaly-Instruct-125k, which presents much higher accuracy and more balanced precision and recall than its original version. This demonstrates the effectiveness of our instruction-tuning dataset. By integrating the anomaly expert with the base model, our Anomaly-OV-0.5B achieves $0.08$ accuracy and $0.06$ F1-score improvements in text-based anomaly detection and better reasoning capability in low-level and complex settings. Equipped with a larger language model, Anomaly-OV-7B provides the best detection performance among all the existing MLLMs and shows comparable reasoning ability with GPT-4o. Notably, we observe that the anomaly expert restricts the detection performance of Anomaly-OV. Therefore, the design of a stronger anomaly expert is suggested for future works.

\begin{table}[ht]
  \begin{minipage}{0.99\linewidth}
\centering
\scalebox{0.80}{
\begin{tabular}{l p{7.5cm} }
\toprule
 \multicolumn{2}{l}{\bf Macaroni Example: Yellowish Spot}  \\
\midrule
&  \includegraphics[height=2.5cm]{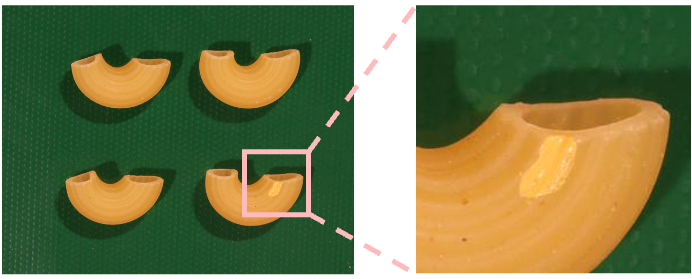} \\

User & Do you see any anomalies in the image? \\
\midrule
GPT-4o & The image shows four pieces of elbow macaroni on a green background. \textcolor{customgreen}{The anomaly is that the macaroni pieces are not whole; they are cut in half.} \\ \midrule
LLaVA-OV & The image shows four pieces of pasta, specifically macaroni shells, arranged on a green textured surface. The pasta appears to be uniformly colored and shaped, with no visible defects or anomalies. \\
\midrule
Ours & Yes, there is an anomaly in the image. \textcolor{customblue}{The bottom right pasta piece has a noticeable yellowish discoloration on its surface.} \\
\bottomrule
\end{tabular}
}
\captionof{table}{Anomaly-OV presents more precise anomaly reasoning.}
\label{tab:pasta}  
  \end{minipage}
  \vspace{-5mm}
\end{table}

\subsection{Extension}
With the generalization and multi-image processing capabilities of MLLMs, it is possible to build a unified assistant for visual inspection. Table \ref{tab:ext} demonstrates the comprehensive knowledge of Anomaly-OV (without using Anomaly-ShapeNet \cite{anomaly_shapenet} for training) on 3D and medical (testing set of BMAD \cite{bmad}) AD \& reasoning. More data, benchmarks, and investigation on a unified model are meaningful.

\begin{table}[ht]
  \begin{minipage}{0.99\linewidth}
\centering
\scalebox{0.80}{
\begin{tabular}{l p{7.5cm} }
\toprule
 \multicolumn{2}{l}{\bf 3D Example: Bulge }  \\
\midrule
&  \includegraphics[height=2.5cm]{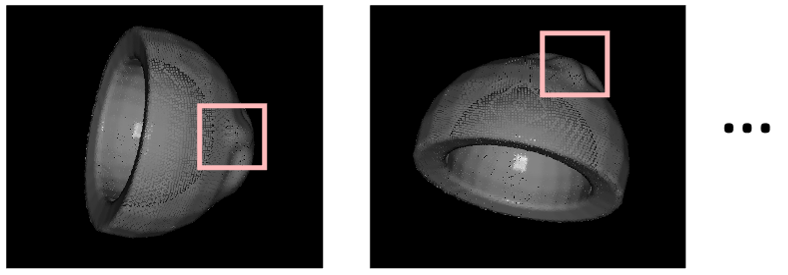} \\

User & Is there any anomaly present in the bowl? \\
\midrule
Anomaly-OV & Yes, \textcolor{customblue}{there is a bulge on the side of the bowl}, which disrupts its otherwise smooth and symmetrical surface. \\
\midrule
 \multicolumn{2}{l}{\bf Medical Example: Pneumonia}  \\
\midrule
&  \includegraphics[height=2.5cm]{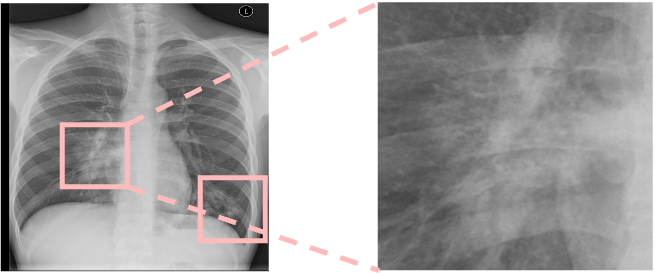} \\

User & What is the most likely diagnosis for this patient? \\
\midrule
Anomaly-OV & Based on the chest X-ray, the most likely diagnosis for this patient is pneumonia. \textcolor{customblue}{The presence of bilateral patchy opacities in the lower lung fields is a common finding in pneumonia}, which is an infection that causes inflammation in the air sacs of the lungs. \\
\midrule
User & What should the patient do next? \\
\midrule
Anomaly-OV & The patient should consult a healthcare professional for further evaluation and appropriate treatment. This may include antibiotics to treat the infection, supportive care such as oxygen therapy, and monitoring of the patient's condition. \\
\bottomrule
\end{tabular}
}
\captionof{table}{Extension to 3D and medical AD \& reasoning.}
\label{tab:ext}  
  \end{minipage}
 \vspace{-5mm}
\end{table}
\section{Conclusion}
In this paper, we establish the first large-scale visual instruction tuning dataset, Anomaly-Instruct-125k, and the corresponding benchmark, VisA-D\&R, to address the data scarcity issue for visual anomaly detection and reasoning. Then, a specialist MLLM, Anomaly-OV, targeting visual inspection is introduced to serve as the baseline in this domain. Anomaly-OV leverages an anomaly expert to assist the LLM with suspicious visual token selection and presents significant improvements on both traditional ZSAD and text-based anomaly detection and reasoning tasks over existing methods. Extension to 3D and medical domains is demonstrated.

{
    \small
    \bibliographystyle{ieeenat_fullname}
    \bibliography{main}
}
\clearpage
\setcounter{section}{0}
\renewcommand\thesection{A\arabic{section}}

\section{Dataset Establishment}
\label{sup_dataset}

\subsection{How to highlight the anomaly?}
\begin{table}[ht]
  \begin{minipage}{0.99\linewidth}
\centering
\scalebox{0.80}{
\begin{tabular}{l p{7.5cm} }
\toprule
 \multicolumn{2}{l}{\bf Prompt Example: Bubble Inside the Capsule}  \\
\midrule
&  \includegraphics[height=2.2cm]{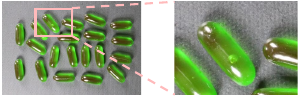} \\

User & Do you observe any anomalies in the image? \\
\midrule
GPT-4o & The image shows green gel capsules that appear uniform in size, shape, and color. Upon inspection, \textcolor{customgreen}{no obvious anomalies such as broken capsules, irregular shapes, or unexpected colors are visible.} \\ \midrule
 \multicolumn{2}{c}{Let's draw a bounding box around the anomaly!} \\ \midrule
&  \includegraphics[height=2.2cm]{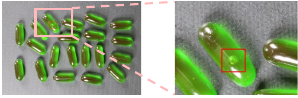} \\

User & Do you observe any anomalies in the image? \\
\midrule
GPT-4o & Yes, there appears to be an anomaly in the image. One of the capsules, marked with a red box, has a noticeable defect. \textcolor{customblue}{It seems to have a bubble or an air pocket inside, which is not observed in the other capsules.} This could indicate a manufacturing defect or inconsistency in the capsule filling process. \\\midrule
 \multicolumn{2}{c}{If there is no bounding box, let's give a short hint to GPT-4o!} \\ \midrule
User & Do you observe any anomalies in the image? (hint: a bubble in the capsule) \\
\midrule
GPT-4o & The anomaly in the image is the presence of a visible air bubble inside one of the green gel capsules. \textcolor{customblue}{While most capsules appear uniformly filled with liquid, the capsule \textbf{located near the top center of the image} has a small, clear air pocket within it.} This could be a result of incomplete filling during the manufacturing process or an issue with the capsule sealing. \\
\bottomrule
\end{tabular}
}
\captionof{table}{Comparison of the GPT-4o \cite{gpt-api-4o} outputs with and without visual and textual hints for the anomaly.}
\label{tab:sup1}  
  \end{minipage}
  \vspace{-2mm}
\end{table}
\noindent
As shown in Table \ref{tab:sup1}, recent advanced MLLMs like GPT-4o fail to detect the anomalies in the image, so building the instruction tuning dataset using previous methods \cite{sharegpt4v} is impractical. However, we observe that when the GPT-4o is provided some "hints", it presents impressive performance on anomaly reasoning or description. For example, a red bounding box drawn around the anomalous area enables GPT-4o to detect the tiny bubble inside the small capsule. This observation indicates that \textcolor{lightpink}{\textbf{the anomaly information is already contained in the visual tokens, and the failure of existing MLLMs is because the language model cannot effectively pick out the related tokens,}} which is the major inspiration of our token-picking mechanism.

Most of the existing AD datasets, such as MVTec AD \cite{mvtec}, contain anomaly masks for anomaly localization. Therefore, we leverage these masks to generate the bounding boxes on the images. Specifically, the masks for an anomalous image are dilated and merged (if two masks are too close) before calculating the coordinates of the bounding boxes. Similarly, the image with bounding boxes drawn on it will serve as the visual prompt for GPT-4o. We also tried many other ways to utilize the anomaly masks, such as highlighting the mask area with different colors, consecutively providing the image and mask, and converting the normalized coordinates of the bounding box into a text prompt. None of them can as effectively guide the GPT-4o in finding anomalous features as drawing bounding boxes on the image.

\begin{figure*}[!t]
\centering
    \includegraphics[width=\textwidth]{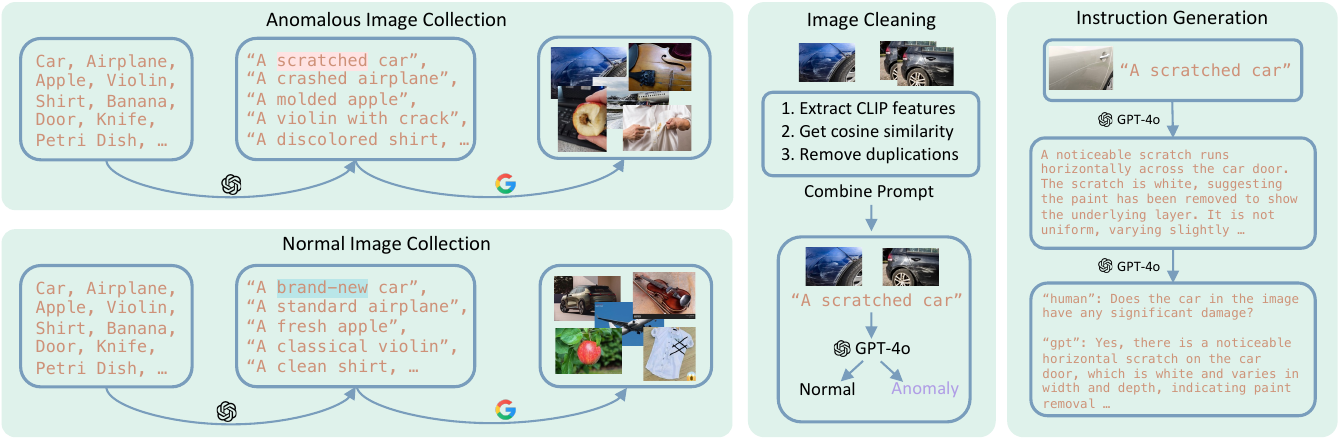}
\caption{Automatic data collection pipeline for WebAD. The entire pipeline is fully automatic at an affordable cost (API usage). Other advanced open-sourced MLLMs can applied to replace GPT-4o for further reduction of cost.}
\label{fig:data_collect}
\end{figure*}

\subsection{WebAD -- The largest AD dataset}
Existing industrial or medical anomaly detection datasets, such as MVTec AD \cite{mvtec} and BMAD \cite{bmad}, only contain a limited number of classes ($<20$) and several different anomaly types for each class (most of the anomaly types are similar) due to the collection of these kinds of anomaly images involves extensive human involvements. This limitation hinders the ZSAD model from learning a generic description of anomaly and normal patterns. Also, the MLLMs cannot obtain enough knowledge of visual anomaly descriptions for unseen anomaly types. Therefore, more diverse data is required for a robust ZSAD \& reasoning model. Many recent dataset works collect and annotate online images to enrich existing datasets and demonstrate their effectiveness in the training of current data-hungry deep learning models. 

To collect the online images that can be utilized for anomaly detection, we design an automatic data collection pipeline by combining GPT-4o \cite{gpt-api-4o} and Google Image Search \cite{google-image-search}. As shown in Figure \ref{fig:data_collect}, we first employ GPT-4o to list 400 class names commonly seen in our daily life. Then, for each class, the GPT-4o is asked to generate 10 corresponding anomalous and normal phrases based on the class name. The abnormality or normality descriptions indicated by these phrases are specifically suitable for the class name. These phrases will serve as the search prompts to query the image links in Google Image Search. However, the downloaded images are very "dirty" and contain many noise samples and duplications. For example, the collected anomaly set contains lots of normal images, and vice versa. A data-cleaning step is applied after the image collection.

\begin{figure*}[!t]
\centering
    \includegraphics[width=\textwidth]{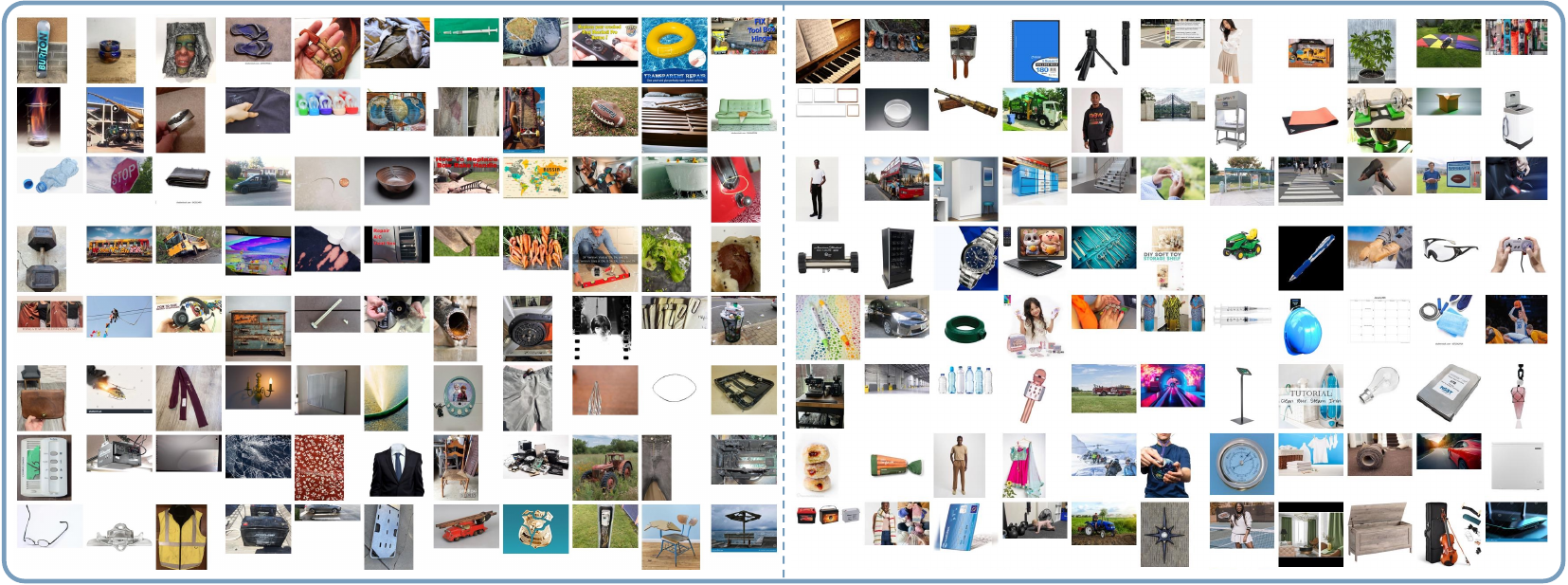}
\caption{Overview of the gallery for in-the-wild image samples in WebAD. The images on the left side are anomalous, while the right side is for normal images. The links to download these images will be released to avoid copyright issues.}
\label{fig:gallery}
\end{figure*}

Since the duplications mainly occur within a specific class, we extract the CLIP \cite{clip} features for all the images in the class and compare the cosine similarity of these features. If the similarity value is larger than $0.99$, then one of the images will be removed. To deal with the problematic grouping of anomaly and normal images, we combine the image and its corresponding search prompt and give them to GPT-4o for normal and anomaly classification. In the system prompt, we explicitly tell the GPT-4o that the search prompt is just a hint and not always correct and ask GPT-4o to determine the normality and abnormality by itself. This step will remove the images with incorrect labels and the artificial images, such as cartons or art. Some samples in the collected WebAD dataset are shown in Figure \ref{fig:gallery}. In total, WebAD contains around 72k images from 380 classes and more than 5 anomaly types for each class.

\subsection{Instruction Data Generation}
\begin{figure*}[!t]
\centering
    \includegraphics[width=\textwidth]{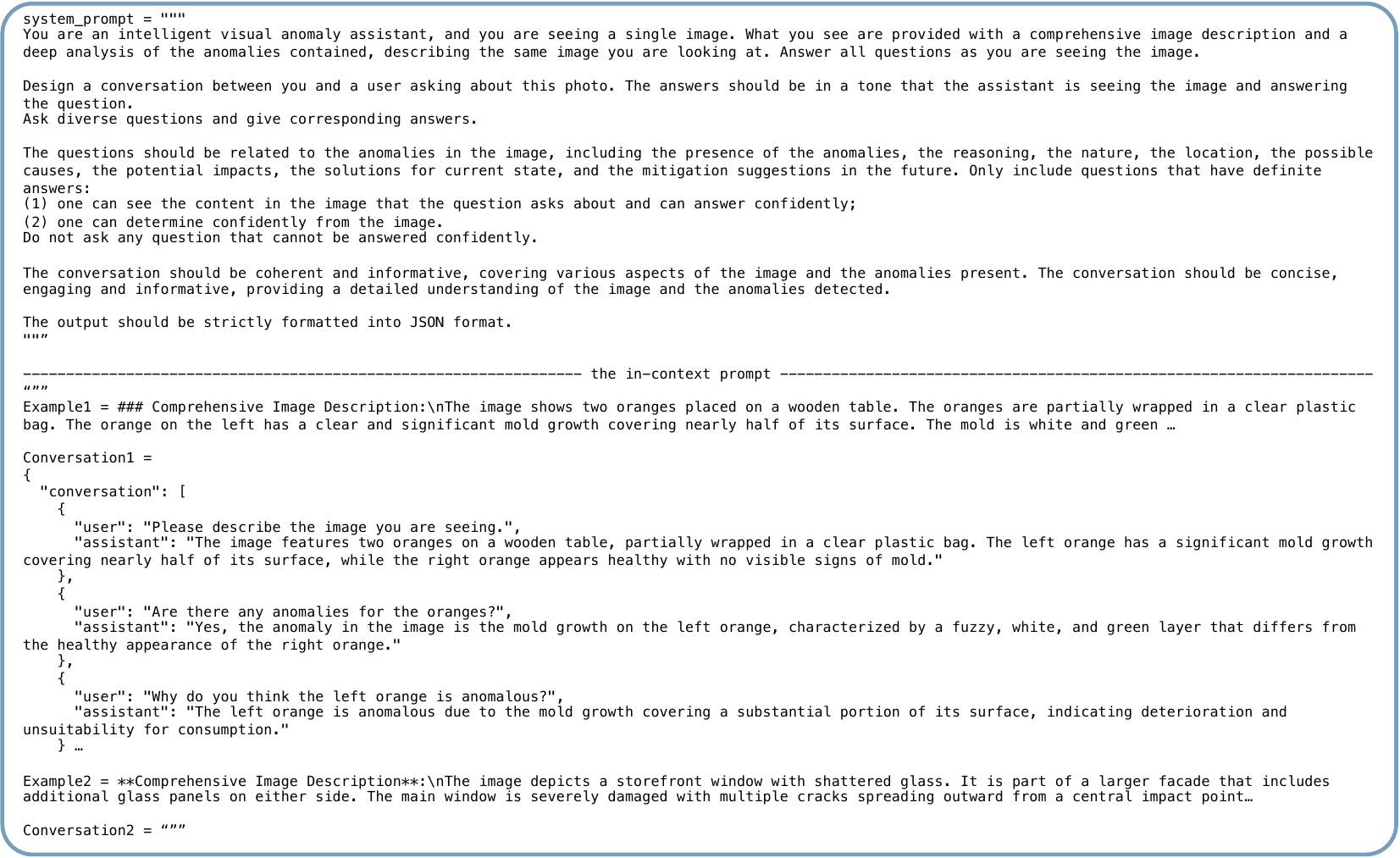}
\caption{Prompt template for generating multi-round conversation in Anomaly-Instruct-125k (modified from the template of LLaVA \cite{llava}).}
\label{fig:in_context}
\vspace{-3mm}
\end{figure*}

For existing datasets, we manually combine the anomaly type and the class name to create the short anomaly prompt (hint). Then, the image with or without the bounding boxes and the corresponding short prompt are utilized to prompt GPT-4o for the generation of detailed descriptions of the image and the anomalies. These descriptions contain all the information required for instruction-following data. The in-context learning strategy is implemented to generate the multi-round conversation data (see Figure \ref{fig:in_context}). Questions designed to elicit a one-word answer are utilized to balance the distribution of the normal and anomaly samples.

\section{Training Details}
In the professional training stage, we leverage AdamW \cite{adamw} to be the optimizer and CosineAnnealingWarmRestarts \cite{loshchilov2017sgdr} as the learning rate scheduler. The initial learning rate is set to be $1e-4$, and the restart iteration is half of the single epoch. The anomaly expert is trained on 8 H100 GPUs for 2 epochs (2 hours), and the total batch size is 128. In the instruction tuning stage, we follow the default training setting of \textit{LLaVA-OneVision} \cite{llavaonevision} (reduce the batch size to 128), and the total training time for 0.5B and 7B models are 7 hours and 50 hours on 8 H100, respectively. When sampling the instruction data from the original recipe of \textit{LLaVA-OneVision}, we put more emphasis on low-level image understanding and 3D multi-view Q\&A, considering that anomaly detection originates from the low-level feature differences and the 3D anomaly detection requires multi-image understanding. Besides, for more knowledge in the medical domain, the model is also fed with the data from LLaVA-Med \cite{llavamed}.

\section{Experimental Results}

\subsection{Anomaly Detection}
\begin{table*}
\small
\centering
\begin{tabular}{l|ccccccccc}
\toprule
VisA     & capsules & fryum      & cashew & macaroni1 & macaroni2 & candle     & pipe fryum & chewinggum & pcb1     \\ 
AUROC    & 98.6     & 93.4       & 93.9   & 91.1      & 70.3      & 96.5       & 99.0       & 96.4       & 90.3     \\
         & pcb2     & pcb3       & pcb4   &           &           &            &            &            &           \\
         & 87.7     & 86.6       & 90.0   &           &           &            &            &            &           \\
\hline
MVTec AD & bottle   & cable      & grid   & capsule   & tile      & pill       & screw      & transistor & hazelnut \\
AUROC    & 95.6     & 88.8       & 99.2   & 92.2      & 95.5      & 92.3       & 98.0       & 88.0       & 91.2     \\
         & carpet   & metal nut  & wood   & zipper    & leather   & toothbrush &            &            &          \\
         & 98.9     & 77.7       & 100.0  & 99.6      & 100.0     & 94.2       &            &            &          \\
\bottomrule
\end{tabular}
\caption{Per-class image-level AUROC of the anomaly expert of Anomaly-OV on VisA and MVTec AD.}
\label{Tab:mvtec_visa}
\vspace{-3mm}
\end{table*}
Similar to previous ZSAD works, the detailed image-level AUROC results for the anomaly expert of Anomaly-OV on VisA \cite{visa} and MVTec AD \cite{mvtec} are provided in Table \ref{Tab:mvtec_visa}.

\subsection{Anomaly Reasoning}
\begin{table}
  \begin{minipage}{0.99\linewidth}
\centering
\scalebox{0.80}{
\begin{tabular}{l p{7.5cm} }
\toprule
 \multicolumn{2}{l}{\bf PCB Example: Bent Transparent LED }  \\
\midrule
&  \includegraphics[height=2.5cm]{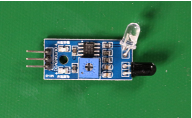} \\

User & Is there an anomaly in the image? \\
\midrule
GPT-4o & The image shows an infrared sensor module, which appears to be intact and properly assembled. There doesn't seem to be any visible anomaly or damage to the components. \\ \midrule
LLaVA-OV & The image shows an electronic circuit board with various components, including a red LED that is not illuminated. There doesn't appear to be anything obviously anomalous in the image. \\
\midrule
Ours & Yes, the anomaly in the image is \textcolor{customblue}{the bent LED on the circuit board, which is not aligned properly with the board's surface.} \\
\bottomrule
\end{tabular}
}
\captionof{table}{Additional results on VisA-D\&R (PCB).}
\label{tab:pcb1}  
  \end{minipage}
\end{table}
\begin{table}
  \begin{minipage}{0.99\linewidth}
\centering
\scalebox{0.80}{
\begin{tabular}{l p{7.5cm} }
\toprule
 \multicolumn{2}{l}{\bf Candle Example: Pink Discoloration}  \\
\midrule
&  \includegraphics[height=2.5cm]{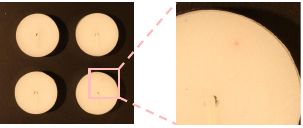} \\

User & Is there an anomaly in the image I should know about? \\
\midrule
GPT-4o & Yes, there is an anomaly in the image. \textcolor{customgreen}{The candle in the top left corner has its wick facing the opposite direction compared to the other three candles.} \\ \midrule
LLaVA-OV & Based on the image, there is no apparent anomaly among the candles. They are all similar in size and shape, with a consistent hole in the center for the wick. The coloration appears uniform across the candles, and there are no visible signs of melting or irregularities that would suggest an anomaly. \\
\midrule
Ours & Yes, the anomaly is \textcolor{customblue}{the presence of a small red spot on the bottom right candle, which contrasts with the otherwise uniform white color.} \\
\bottomrule
\end{tabular}
}
\captionof{table}{Additional results on VisA-D\&R (Candle).}
\label{tab:candle}  
  \end{minipage}
\end{table}
\begin{table}[t]
  \begin{minipage}{0.99\linewidth}
\centering
\scalebox{0.80}{
\begin{tabular}{l p{7.5cm} }
\toprule
 \multicolumn{2}{l}{\bf Capsule Example: Dark Spot }  \\
\midrule
&  \includegraphics[height=2.5cm]{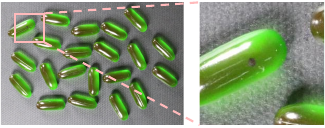} \\

User & Is there an anomaly present in the image? \\
\midrule
GPT-4o & Yes, there is an anomaly in the image. \textcolor{customgreen}{One of the capsules is a different color; it is brown instead of green.} \\ \midrule
LLaVA-OV & Based on the image alone, there is no clear indication of an anomaly. The pills appear to be uniformly shaped and colored, with no visible defects or irregularities. However, without additional context or a point of reference for size, it's difficult to determine if there are any anomalies that might not be immediately apparent from the image.  \\
\midrule
Ours & Yes, there is an anomaly present in the image. \textcolor{customblue}{One of the capsules has a darker spot on its surface, which contrasts with the otherwise uniform green color.} \\
\bottomrule
\end{tabular}
}
\captionof{table}{Additional results on VisA-D\&R (Capsules).}
\label{tab:capsule}  
  \end{minipage}
  \vspace{-1mm}
\end{table}
\begin{table}[ht]
  \begin{minipage}{0.99\linewidth}
\centering
\scalebox{0.80}{
\begin{tabular}{l p{7.5cm} }
\toprule
 \multicolumn{2}{l}{\bf Fryum Example: Break Edge }  \\
\midrule
&  \includegraphics[height=2.5cm]{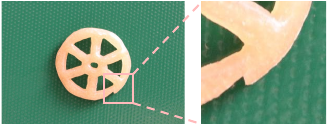} \\

User & Is there any anomaly in the image? \\
\midrule
GPT-4o & The image shows a wheel-shaped snack on a green surface. There doesn't appear to be any obvious anomaly. \\ \midrule
LLaVA-OV & The image shows a circular object with a pattern that resembles a wheel or a gear, placed on a green textured surface. There doesn't appear to be any obvious anomaly in the image; it seems to be a simple photograph of an object against a background. If there is something specific you are referring to as an anomaly, please provide more details or clarify what aspect you are questioning.  \\
\midrule
Ours & Yes, there is an anomaly in the image. The pasta wheel has \textcolor{customblue}{a noticeable crack on its lower right side, disrupting its otherwise smooth and continuous surface.} \\
\bottomrule
\end{tabular}
}
\captionof{table}{Additional results on VisA-D\&R (Fryum).}
\label{tab:fryum}  
  \end{minipage}
\end{table}
\begin{table}
  \begin{minipage}{0.99\linewidth}
\centering
\scalebox{0.80}{
\begin{tabular}{l p{7.5cm} }
\toprule
 \multicolumn{2}{l}{\bf Cashew Example: Brown Spot }  \\
\midrule
&  \includegraphics[height=2.5cm]{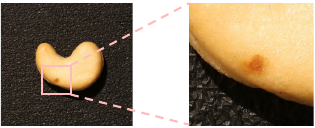} \\

User & Does the image contain any anomaly? \\
\midrule
GPT-4o & The image shows a cashew nut on a textured black surface. There are no visible anomalies; the cashew appears normal. \\ \midrule
LLaVA-OV & Based on the image provided, there are no visible anomalies. The object appears to be a single, intact cashew nut with a smooth surface and a natural brown coloration that is typical for cashews. There are no signs of damage, such as cracks or holes, and the background does not show any irregularities that would suggest an anomaly in the context of the image.  \\
\midrule
Ours & Yes, there is a noticeable anomaly on the cashew nut: \textcolor{customblue}{a small brown spot located near the center of the nut.} \\
\bottomrule
\end{tabular}
}
\captionof{table}{Additional results on VisA-D\&R (Cashew).}
\label{tab:cashew}  
  \end{minipage}
\end{table}
\begin{table}
  \begin{minipage}{0.99\linewidth}
\centering
\scalebox{0.80}{
\begin{tabular}{l p{7.5cm} }
\toprule
 \multicolumn{2}{l}{\bf In-the-Wild Example: Graffitied Road Sign }  \\
\midrule
&  \includegraphics[height=3.5cm]{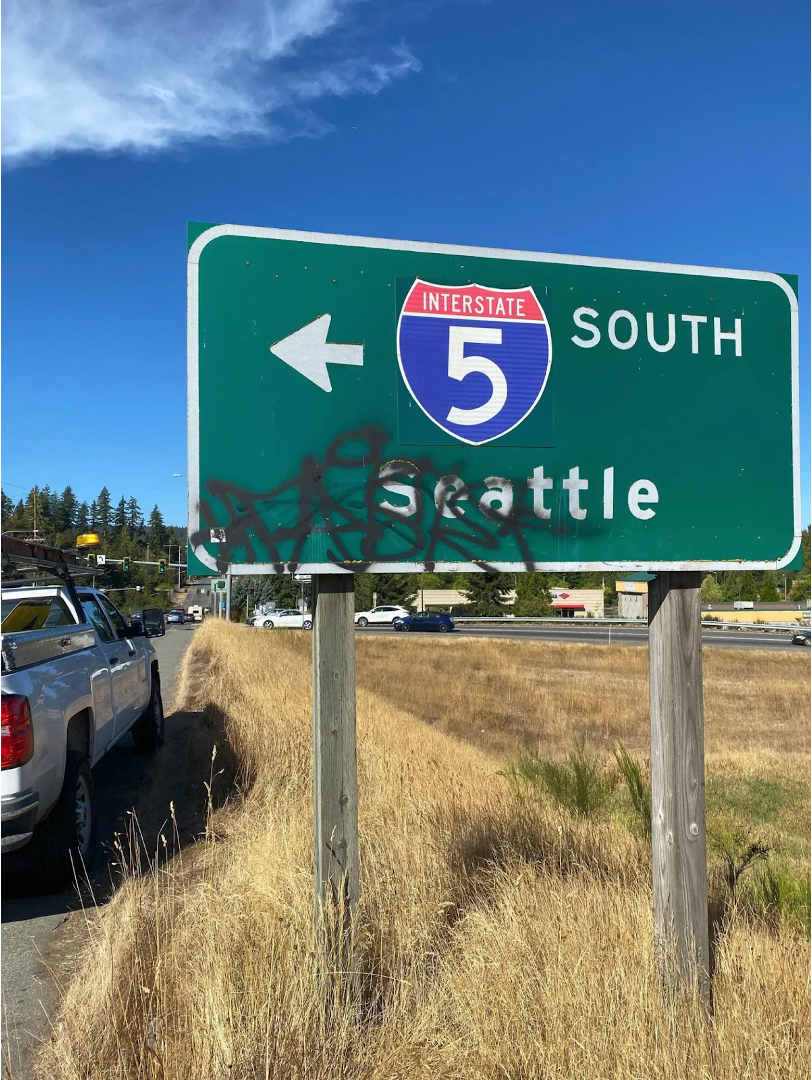} \\

User & Describe the anomaly in the image. \\
\midrule
Anomaly-OV & The anomaly in the image is the graffiti on the highway sign, which includes black spray paint marks and scribbles over the word 'Seattle' and other parts of the sign. \\ 
\midrule
User & What is the reason for the anomaly?  \\
\midrule
Anomaly-OV & The graffiti is likely the result of vandalism, where individuals have used spray paint to deface public property for personal expression or as a form of protest. \\
\midrule
User & How does the anomaly affect the functionality of the object?  \\
\midrule
Anomaly-OV & The graffiti affects the sign's functionality by potentially obstructing the visibility of important information, such as the word 'Seattle,' which could confuse drivers and lead to navigational errors. \\
\midrule
User & How to avoid the anomaly in the future?  \\
\midrule
Anomaly-OV & To avoid such anomalies, regular maintenance and cleaning of signs should be implemented. Additionally, community awareness programs can help reduce vandalism by promoting respect for public property. \\
\bottomrule
\end{tabular}
}
\captionof{table}{In-the-wild results for an unseen object (Road Sign).}
\label{tab:road_sign}  
  \end{minipage}
\end{table}
Table \ref{tab:pcb1} to \ref{tab:cashew} presents more comparison results of GPT-4o \cite{gpt-api-4o}, \textit{LLaVA-OneVision} \cite{llavaonevision}, and Anomaly-OV on AD \& reasoning. Anomaly-OV shows better performance in the detection and description of the visual anomalies in the images. Table \ref{tab:road_sign} demonstrates the low-level and complex reasoning capability of Anomaly-OV for an in-the-wild image, indicating a comprehensive understanding of the anomaly.

\section{Limitation and Future Work}
\begin{table}
  \begin{minipage}{0.99\linewidth}
\centering
\scalebox{0.80}{
\begin{tabular}{l p{7.5cm} }
\toprule

\multicolumn{2}{l}{\bf Failure Example 1: Wrong Classification}  \\
\midrule
&  \includegraphics[height=2.5cm]{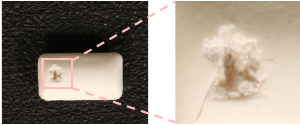} \\
User & Is there any anomaly visible in the image? \\
\midrule
Anomaly-OV & Yes, there is a noticeable anomaly on the \textcolor{customgreen}{soap bar}: a hole with jagged edges located towards the left side. \\
\midrule
 \multicolumn{2}{l}{\bf Failure Example 2: General Description }  \\
\midrule
&  \includegraphics[height=2.5cm]{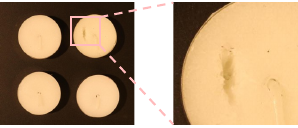} \\

User & Are there any anomalies present in the image with the set of candles? \\
\midrule
Anomaly-OV & Yes, there is an anomaly present in the top right candle. It has a noticeable \textcolor{customgreen}{crack} on its surface, which is not typical for a smooth, intact candle. \\
\midrule
 \multicolumn{2}{l}{\bf Failure Example 3: Wrong Reasoning }  \\
\midrule
&  \includegraphics[height=2.5cm]{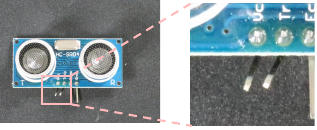} \\

User & Are there any anomalies in the image? \\
\midrule
Anomaly-OV & Yes, the anomaly in the image is \textcolor{customgreen}{the absence of a crystal oscillator on the HC-SR04 ultrasonic sensor module.} \\

\bottomrule
\end{tabular}
}
\captionof{table}{Failure results of Anomaly-OV on VisA-D\&R.}
\label{tab:failure}  
  \end{minipage}
\end{table}
\noindent
\textbf{Limitation.} As shown in Table \ref{tab:failure}, sometimes, Anomaly-OV fails to provide an accurate classification of the target object, describes the anomaly by a general word (wax missing is described by "crack"), or presents wrong reasoning with hallucination. Also, there is still a large space for improvement in the detection performance of Anomaly-OV. Besides, the images contained in VisA-D\&R  are from the industrial domain, so more benchmarks in other domains, such as 3D and medical anomaly detection, are required to evaluate a unified AD \& reasoning model.

\medskip

\noindent
\textbf{Future Work.} The detection performance of Anomaly-OV is highly determined by the anomaly expert (see Table \ref{Tab:3}), so a more advanced design of the expert model is recommended in future research. One can change the base model to other open-sourced MLLMs to resolve the wrong classification issue. Also, we found that the diversity of the anomaly type is very limited in existing industrial anomaly datasets (mainly 'crack' or 'broken'), causing the assistant to fail to provide fine-grained anomaly reasoning or description for unseen anomaly features. Therefore, a more diverse industrial anomaly detection dataset is urgently required. Similar to other traditional MLLMs, Anomaly-OV only utilizes the output visual tokens from the last layer of the visual encoder as the input for LLM. However, anomaly detection is highly dependent on low-level visual clues. Hence, \textbf{forwarding multi-level features from different layers to the LLM} (as recent paper: "Dense Connector for MLLMs" \cite{yao2024denseconnectormllms} ) should be a possible solution for performance improvement.


\end{document}